# Distributed Linguistic Representations in Decision Making: Taxonomy, Key Elements and Applications, and Challenges in Data Science and Explainable Artificial Intelligence

Yuzhu Wu[a], Zhen Zhang[b], Gang Kou[a], Hengjie Zhang[c], Xiangrui Chao[d],

Cong-Cong Li[e], Yucheng Dong[d,f,*], Francisco Herrera[g,h]

[a] School of Business Administration, Southwestern University of Finance and Economics, Chengdu 611130, China
[b] Institute of Systems Engineering, Dalian University of Technology, Dalian 116024, China
[c] Business School, Hohai University, Nanjing 211100, China
[d] Center for Network Big Data and Decision-Making, Business School, Sichuan University, Chengdu 610065, China
[e] School of Economics and Management, Southwest Jiaotong University, Chengdu 610031, China
[f] School of Information Management and Statistics, Hubei University of Economics, Wuhan 430205, China
[g] Andalusian Research Institute on Data Science and Computational Intelligence (DaSCI), University of Granada, Granada 18071, Spain
[h] Faculty of Computing and Information Technology, King Abdulaziz University, Jeddah, Saudi Arabia

**Abstract:** Distributed linguistic representations are powerful tools for modelling the uncertainty and complexity of preference information in linguistic decision making. To provide a comprehensive perspective on the development of distributed linguistic representations in decision making, we present the taxonomy of existing distributed linguistic representations. Then, we review the key elements and applications of distributed linguistic information processing in decision making, including the distance measurement, aggregation methods, distributed linguistic preference relations, and distributed linguistic multiple attribute decision making models. Next, we provide a discussion on ongoing challenges and future research directions from the perspective of data science and explainable artificial intelligence.

## 1. Introduction

Computing with words (CW) models [29], [56], [80], [94], [101] play a key role in dealing with information linguistically. The classical CW models emphasize on handling linguistic information based on the use of membership functions [101]-[104]. In CW, a mainstream is the linguistic symbolic computational models [14], [94]-[96], among which the 2-tuple linguistic representation model [31] is widely used to handle balanced linguistic information in the form of linguistic 2-tuples. Thereafter, different methods are proposed to process unbalanced linguistic information, including the proportional 2-tuple linguistic model [82], unbalanced linguistic term set [30], numerical scale [18], and nonuniform ordered qualitative scale [23]. Considering individuality of decision makers in expressing and understanding words, personalized individual semantics (PIS) based models [16], [38]-[41] are introduced following the stream of linguistic symbolic computational models. Another mainstream are the type-2 fuzzy sets, because words mean different things to different people [60], the type-2 fuzzy sets are used to deal with this issue [56]-[59].

* Corresponding author.
Email addresses: wuyuzhu@swufe.edu.cn (Y. Wu), zhen.zhang@dlut.edu.cn (Z. Zhang), kougang@swufe.edu.cn (G. Kou), hengjiezhang@hhu.edu.cn (H. Zhang), chaoxr@scu.edu.cn (X. Chao), congcongli@swjtu.edu.cn (C. Li), ycdong@scu.edu.cn (Y. Dong), herrera@decsai.ugr.es (F. Herrera).

In the last decade has been proposed two different concepts for representing cognitive complex information, in 2012 the hesitant fuzzy linguistic term set (HFLTS) [70] and in 2014 the linguistic distribution (LD) [107], becoming two approaches in modelling hesitant and uncertain linguistic information in decision making. (i) HFLTSs are well qualified to represent decision makers' hesitant preferences by using comparative linguistic expressions. (ii) LDs provide certain symbolic proportion information over linguistic terms to describe distributed preferences of decision makers as distributed assessments. In [66], Pang et al. introduced the probabilistic linguistic term set (PLTS), with a different name for the similar concept, and the use of LD.

Compared with simple linguistic 2-tuples, complex expressions like HFLTS and LD have improved the flexibility of expression and have shown to be very useful to deal with complex decision making problems. In practical decision contexts with complexity, decision makers are often uncertain and hesitant to make decisions due to tight time pressure and lack of knowledge, and their linguistic preference information may be presented in the form of distributed linguistic representations [17], [107]. These phenomena have raised the rapidly growing demand of modelling and processing distributed preferences with efficiency, and advanced distributed linguistic representations to become increasingly popular in decision making.

The aim of this paper is to present a review of the current hot topics of distributed linguistic representations in decision making, including the taxonomy of distributed linguistic representations, the key elements in distributed linguistic decision making, and some challenges and future research directions from the perspective of data science and explainable artificial intelligence (XAI).

The remainder of this paper is organized as follows. Section 2 introduces the origin, basic concepts and taxonomy of existing distributed linguistic representations. Then Section 3 presents the key elements and applications of distributed linguistic representations in decision making. Next, the challenges and future research directions are discussed in Section 4. Finally, the main conclusions are drawn in Section 5.

## 2. Distributed linguistic representations: origin, basic concepts, and taxonomy

In this section, we present the origin, basic concepts, and taxonomy of distributed linguistic representations.

### 2.1 Origin of distributed linguistic representation: linguistic 2-tuples and HFLTS

In the development of linguistic decision making, various distributed linguistic representations have been proposed to model decision makers' linguistic preferences. These distributed linguistic representations are basically derived from the following two types of basic linguistic expression formats:

(1) Linguistic 2-tuples, which show their convenience in preferences construction via single or two successive linguistic terms;

(2) HFLTS, which is well qualified to elicit hesitant preferences by using several consecutive linguistic terms.

This section reviews the basic concepts of linguistic 2-tuples and HFLTS.

**Definition 1 (2-tuple linguistic representation model) [31], [55]**: Let $L = \{l_0, l_1, ..., l_g\}$ be an established linguistic term set with odd cardinality satisfying the required characteristics: (i) The set is ordered: $l_i > l_j$ if $i > j$; (ii) A negation operator: $Neg(l_i) = l_j$ such that $j = g - i$, and g+1 is the cardinality of $L$. Let $\beta \in [0, g]$ be a numerical value representing the result of a symbolic aggregation operation. Then, a linguistic 2-tuple $(l_i, \alpha)$, where $l_i \in L$ and $\alpha \in [-0.5, 0.5)$ that expresses the equivalent information to $\beta$ is obtained by means of a one to one mapping: $\Delta : [0, g] \to L \times [-0.5, 0.5)$, and

$$\Delta(\beta) = (l_i, \alpha) \qquad (1)$$

with $\begin{cases} l_i, i = round(\beta) \\ \alpha = \beta - i, \alpha \in [-0.5, 0.5) \end{cases}$ where *round* is the usual round operation.

Let $\bar{L} = \{(l_i, \alpha) \mid l_i \in L, \alpha \in [-0.5, 0.5)\}$ be the set of all the linguistic 2-tuples associated to $L$. The inverse function of $\Delta$ is denoted by: $\Delta^{-1} : \bar{L} \to [0, g]$ with $\Delta^{-1}(l_i, \alpha) = i + \alpha$. If $\alpha = 0$, the linguistic 2-tuple $(l_i, \alpha)$ is a simple term $l_i$. And the negation operator of a linguistic 2-tuple $(l_i, \alpha)$ is defined by $Neg(l_i, \alpha) = \Delta(g - (\Delta^{-1}(l_i, \alpha)))$.

Linguistic 2-tuple and its computational model [31] have shown good advantages in constructing simple linguistic preferences of decision makers in linguistic decision making. An overview on the 2-tuple linguistic representation model for CW in decision making can be found in [55].

Wang and Hao [82] proposed the proportional linguistic 2-tuples, which add symbolic proportion information to two successive linguistic terms.

**Definition 2 (Proportional 2-tuple linguistic representation model) [82]**: Let $L=\{l_0,l_1,...,l_g\}$ be an established linguistic term set, $I=[0,1]$ and $IL \equiv I \times L = \{(\gamma_i, l_i) | \gamma_i \in [0,1], l_i \in L\}$. Given a pair $(l_i, l_{i+1})$ of two successive terms of $L$, any two elements $(\gamma_i, l_i)$ and $(\gamma_{i+1}, l_{i+1})$ of $IL$ are called a symbolic proportion pair, and $(\gamma_i, \gamma_{i+1})$ are called a pair of symbolic proportions of the pair $(l_i, l_{i+1})$ if $\gamma_i + \gamma_{i+1} = 1$. A symbolic proportion pair $(\gamma_i, l_i)(1-\gamma_i, l_{i+1})$ can be denoted as $(\gamma_i l_i, (1-\gamma_i)l_{i+1})$, and the proportional 2-tuple set generated by $L$ is denoted by

$$\bar{\bar{L}} = \{(\gamma_i l_i, (1-\gamma_i)l_{i+1}) | \gamma_i \in [0,1], l_i \in L\}. \quad (2)$$

The elements of $\bar{\bar{L}}$ are called linguistic proportional 2-tuples.

The negation operator for proportional linguistic 2-tuples is defined by $Neg((\gamma_i l_i, (1-\gamma_i)l_{i+1})) = ((1-\gamma_i)l_{g-i-1}, \gamma_i l_{g-i})$.

To improve the flexibility of linguistic expressions, HFLTS [70] was proposed, which acts as a tool in fulfilling decision makers' hesitant necessities and requirements by applying comparative linguistic expressions (several consecutive linguistic terms). The concept of HFLTS is introduced as Definition 3.

**Definition 3 (HFLTS) [70]**: Let $L=\{l_0,l_1,...,l_g\}$ be an established linguistic term set. An HFLTS, $h_L$, is an ordered finite subset of consecutive linguistic terms of $L$. The expression of $h_L$ is given by

$$h_L = \{l_i, l_{i+1}, ..., l_j | 0 \leq i \leq j \leq g\}. \quad (3)$$

The negation operation for $h_L$ is defined by $Neg(h_L) = \{l_{g-i} | l_i \in h_L, i \in \{0,1,...,g\}\}$.

An approach to generate HFLTS based preferences via comparative linguistic expressions by using a context-free grammar was presented in [70]. The recent developments for HFLTS in decision making can refer to [69].

The distributed linguistic representations in this review have two origins:

(1) Extension of the proportional linguistic 2-tuple based representation, referring to Zhang et al. [107], Guo et al. [25], [26];

(2) HFLTS with distributed preference information, referring to Wu and Xu [90], Chen et al. [9], Pang et al. [66], Zhang et al. [109], etc.

In particular, distributed linguistic representations can be formed in two ways: linguistic preference expressions with distributed information in an individual context ([17], [107], etc.), and information fusion of linguistic terms or HFLTSs in a group context ([46], [100], [118], [119], etc.), which can be described in Fig. 1.

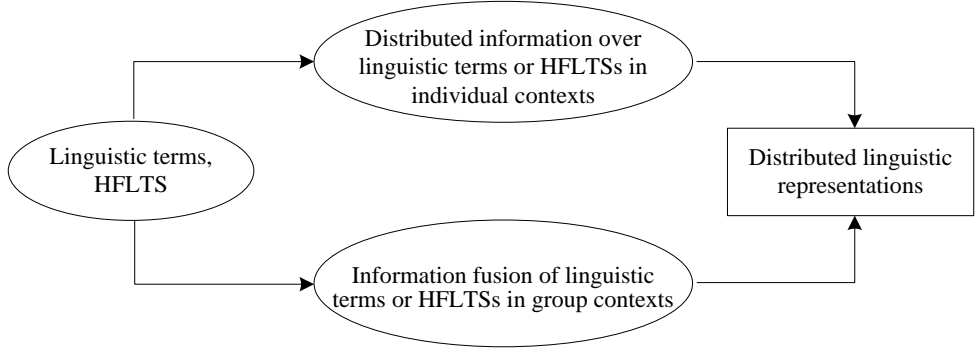

Fig.1 Formation of distributed linguistic representations in decision making

The distributed linguistic representations are reviewed in the following sections in detail.

## 2.2 Linguistic distribution

Compared with HFLTS, LD provides some symbolic proportion information over linguistic terms. The basic concept of LD is presented as Definition 4.

**Definition 4 (LD) [107]**: Let $L=\{l_0,l_1,...,l_g\}$ be an established linguistic term set. An LD over $L$ is defined by

$$D_L = \{(l_i, \rho(l_i)) \mid i=0,1,...,g\}, \qquad (4)$$

where $\rho(l_i) \geq 0$ is the symbolic proportion of $l_i$, and $\sum_{i=0}^{g} \rho(l_i) = 1$.

In an LD $D_L$, $l_i$ represents a linguistic term used by decision makers, and $\rho(l_i)$ represents the associated symbolic proportion information of $l_i$ as a probabilistic distribution associated to the linguistic terms (really it is a complete probabilistic distribution).

The expectation of $D_L$ is defined by: $E(D_L) = \Delta\left(\sum_{i=0}^{g}\left(\Delta^{-1}(l_i) \times \rho(l_i)\right)\right)$, and the negation operator of $D_L$ was proposed [107]: $Neg(D_L) = \{(l_i, \rho(l_{g-i})) \mid (l_i, \rho(l_i)) \in D_L\}$.

## 2.3 Incomplete versions of linguistic distributions

Distribution preference information provided by decision makers is not always complete, and in the following we review the basic concepts of distributed linguistic representations with incompleteness: proportional linguistic distribution (PLD), PLTS, and incomplete linguistic distribution (ILD).

(1) PLD

Based on the proportional 2-tuple linguistic representation model, Guo et al. [26] discussed the LD with partial symbolic information by introducing the concept of PLD through a different mathematical representation.

**Definition 5 (PLD) [26]**: Let $L=\{l_0,l_1,...,l_g\}$ be an established ordinal linguistic term set, and $I=[0,1]$, $IL \equiv I \times L = \{(\gamma_i, l_i) \mid \gamma_i \in [0,1], l_i \in L\}$ be the set of proportional 2-tuples of *L*. Given a sequence of *r*+1 successive ordinal terms of *L*, any *r*+1 elements $(\gamma_i, l_i)$, $(\gamma_{i+1}, l_{i+1})$,…, $(\gamma_{i+r}, l_{i+r})$ of *IL* are called a symbolic proportion sequence. The PLD is defined by

$$D_L^p = \{(\gamma_i l_i, \gamma_{i+1} l_{i+1}, \ldots, \gamma_{i+r} l_{i+r}, \varepsilon) \mid l_i \in L\} \tag{5}$$

where $\gamma_i, \ldots, \gamma_{i+r} \in (0,1]$, $0 < \sum_{j=i}^{i+r} \gamma_j \leq 1$, and $0 \leq i$, $i + r \leq g$. $\varepsilon = 1 - \sum_{j=i}^{i+r} \gamma_j$ represents the extent of incomplete information.

In a PLD $D_L^p$, $l_i$ represents the preference judgment provided by decision makers, and $\gamma_i$ is the proportional coefficient of $l_i$, which represents the decision maker's confidence level that he/she believes a linguistic term fits an evaluation. If $\sum_{j=i}^{i+r} \gamma_j = 1$, $D_L^p$ is called a complete PLD; if $\sum_{j=i}^{i+r} \gamma_j < 1$, $D_L^p$ is called an incomplete PLD.

The set of all the symbolic proportion sequences is called proportional linguistic distribution set, denoted by $L^*$: $L^* = \{(\gamma_i l_i, \gamma_{i+1} l_{i+1}, \ldots, \gamma_{i+r} l_{i+r}, \varepsilon) \mid 0 \leq i, i+r \leq g, l_i \in L\}$.

In [25], Guo et al. discussed the PLD with interval symbolic proportions.

(2) PLTS

Pang et al. [66] proposed the PLTS to solve decision makers' preferences with hesitancy among several possible linguistic terms. Moreover, the incomplete probabilistic distribution information of certain linguistic terms is considered.

**Definition 6 (PLTS) [66]**: Let $L = \{l_0, l_1, \ldots, l_g\}$ be an established ordinal linguistic term set. A PLTS is defined by

$$L(p) = \{l_i(p_i) \mid l_i \in L, p_i \geq 0, i = 1, \ldots, n(L(p))\} \tag{6}$$

where $\sum_{i=0}^{n(L(p))} p_i \leq 1$, and $n(L(p))$ denotes the number of element $l_i$ in $L(p)$.

In a PLTS $L(p)$, $l_i$ represents decision maker's preference and $p_i$ is the probability of $l_i$. If $\sum_{i=0}^{n(L(p))} p_i = 1$, the probability distribution information of $L(p)$ is complete; if $\sum_{i=0}^{n(L(p))} p_i < 1$, partial ignorance exists because of decision makers' knowledge limitation.

A computational model for handling incompleteness of $L(p)$ is presented in [66].

**Remark 1.** A PLTS is transformed into an LD by using the normalization method proposed in [66]. However, such transformation is questionable [40], which can be demonstrated by using the following example. Let $L = \{l_0, l_1, l_2, l_3, l_4\}$ be a linguistic term set and $L(p) = \{l_3(0.3), l_4(0.3)\}$ be a PLTS. By using the normalization method proposed in [66], an LD $D_L = \{(l_3, 0.5), (l_4, 0.5)\}$ can be obtained. Obviously, $L(p) = \{l_3(0.3), l_4(0.3)\}$ and $D_L = \{(l_3, 0.5), (l_4, 0.5)\}$ represent different preference information. In fact, for two PLTSs $L^r(p) = \{(l_i^r(p_i^r) | l_i^r \in L, p_i^r \geq 0, i = 1, 2..., n(L^r(p))\}$ and $L^s(p) = \{(l_i^s(p_i^s) | l_i^s \in L, p_i^s \geq 0, i = 1, 2..., n(L^s(p))\}$, if $n(L^r(p)) = n(L^s(p))$, $l_i^r = l_i^s$, and $p_i^r = \theta p_i^r$ $(i = 1, 2, ..., n(L^r(p)))$, then $L^r(p)$ and $L^s(p)$ will be transformed into the same LD by using the normalization method proposed in [66], which is unreasonable.

(3) ILD

Zhang et al. [105] proposed the ILD, as shown in Definition 7.

**Definition 7 (ILD) [85], [105]**: Let $L = \{l_0, l_1, ..., l_g\}$ be an established ordinal linguistic term set. An ILD of $L$ is given by

$$D_L^I = \{(l_i, \rho(l_i)), \eta \mid i = 0, 1, ..., g\} \tag{7}$$

where $l_i \in L$, $\sum_{i=0}^{g} \rho(l_i) + \eta = 1$, and $\rho(l_i), \eta \in [0,1]$.

An ILD can be used to express a decision maker's preferences with incompleteness. In an ILD $D_L^I$, $l_i$ represents the linguistic term used by decision makers, and $\rho(l_i)$ represents the relevant symbolic proportion information of $l_i$. The variable $\eta$ represents the extent of incompleteness and uncertainty in the preference $D_L^I$. If $\eta = 0$, $D_L^I$ is complete and equivalent to the LD in Definition 4. Otherwise, the larger $\eta$ indicates the greater uncertainty in $D_L^I$.

## 2.4 Flexible linguistic expression

HFLTSs and LDs have provided convenience in the construction of complex linguistic expressions, but as a general format of linguistic preference expressions, flexible linguistic expression (FLE) shows its advantages in linguistic decision making, which is formally defined as follows.

**Definition 8 (FLE) [86], [87]**: Let $L = \{l_0, l_1, ..., l_g\}$ be an established linguistic term set, and $S_L$ be a set whose elements $s_L$ are the subsets of $L$. The decision maker expresses his/her preferences by presenting distribution information of $s_L$. Then the decision maker's preference is an FLE, denoted as $f_L$, and it is formally defined by

$$f_L = \{(s_L, \rho(s_L)) \mid s_L \in S_L\} \tag{8}$$

where $\rho(s_L) \in [0,1]$.

Let the set of all the subsets of $L$ be $S$. The set $S_L$ is not fixed and $S_L \subset S$. In an FLE $f_L$, $s_L$ represents the preference expressed by decision makers, and $\rho(s_L)$ represents the associated symbolic proportion information of $s_L$. If $s_L \in S$ and $s_L \notin S_L$, it means that $s_L$ is not used by the decision maker to represent his/her preference. The negation operator of $f_L$ is defined by [87]:

$Neg(f_L) = \{(Neg(s_L), \rho(s_L)) \mid s_L \in S_L\}$, where $Neg(s_L) = \{l_{g-i} \mid l_i \in s_L\}$.

### 2.5 Other variants of LD

Moreover, there exist several representative distributed linguistic representations, including LD with interval symbolic proportions (INLD) [17], possibility distribution for HFLTS (PDHFLTS) [90], and proportional HFLTS (PHFLTS) [9], and hesitant linguistic distribution (HLD) [109]. These variants are introduced below.

Dong et al. [17] proposed the version of LD with interval symbolic proportions.

**Definition 9 (INLD) [17], [89]**: Let $L = \{l_0, l_1, ..., l_g\}$ be an established linguistic term set. An INLD is given by

$$\tilde{D}_L = \{(l_i, \rho(l_i)) \mid l_i \in L, i = 0, 1, ..., g\} \quad (9)$$

where $\rho(l_i)$ is the interval symbolic proportion of $l_i$ satisfying $\rho(l_i) = [\rho^L(l_i), \rho^U(l_i)] \subseteq [0,1]$. $\rho^L(l_i)$ and $\rho^U(l_i)$ are the lower bound and upper bound of $\rho(l_i)$ respectively.

In an INLD $\tilde{D}_L$, $\rho(l_i)$ represents the associated interval proportion information of linguistic term $l_i$, and the interval length of $\rho(l_i)$ reflects the confidence level of decision maker when providing the preference $l_i$. If $\rho^L(l_i) = \rho^U(l_i)$, $\tilde{D}_L$ is mathematically consistent with the LD in Definition 4.

Wu and Xu [90] proposed the PDHFLTS in group decision making (GDM) context, in which decision makers provide their preferences via single terms and possibility information is distributed over terms in an HFLTS.

**Definition 10 (PDHFLTS) [90]**: Let $L = \{l_0, l_1, ..., l_g\}$ be an established linguistic term set, and $h_L$ be an HFLTS of $L$ given by a decision maker. The PDHFLTS is represented by

$$D_{h_L}^P = \{(l_i, p_i) \mid l_i \in h_L\} \quad (10)$$

where $P = (p_0, p_1, ..., p_g)$ is the possibility distribution of $h_L$, $p_i = \begin{cases} 1/n(h_L) & l_i \in h_L \\ 0 & otherwise \end{cases}$ denotes the possibility assigned over the linguistic term $l_i$, $\sum_{i=0}^{g} p_i = 1$, and $n(h_L)$ is the number of elements in $h_L$.

In a PDHFLTS $D_{h_L}^P$, preference from a decision maker is represented by an HFLTS, and each term $l_i$ in this HFLTS has the same possibility $p_i$ to become the assessment value of an alternative.

Chen et al. [9] proposed the PHFLTS to solve HFLTS based preferences provided by decision makers in GDM problems.

**Definition 11 (PHFLTS) [9]**: Let $L=\{l_0,l_1,...,l_g\}$ be an established linguistic term set, and $h_L^k$ ($k=1,...,n$) be $n$ HFLTSs given by $n$ decision makers. A PHFLTS based on the union of $h_L^k$ is a set of ordered finite proportional pairs, represented by

$$P_{H_L} = \{(l_i, p_i) | l_i \in L\} \tag{11}$$

where $(l_i, p_i)$ is called a proportional linguistic pair, $P=(p_0, p_1,...,p_g)^T$ is a proportional vector and $p_i \in [0,1]$ represents the possibility degree of $l_i$ provided by $n$ decision makers, and $\sum_{i=0}^{g} p_i = 1$.

Chen et al. [9] discussed the case of PHFLTS with incomplete proportion information, and mentioned that PHFLTS is mathematically consistent with LD in Definition 4. Some associated computational approaches can be found in [10]-[12].

Zhang et al. [109] proposed the HLD, in which preferences of decision makers are HFLTSs.

**Definition 12 (HLD) [109]**: Let $L=\{l_0,l_1,...,l_g\}$ be an established linguistic term set, and $h_L$ be an HFLTS of $L$ given by a decision maker. Let the set of all the HFLTS of $L$ be $H_L$. The HLD is defined by

$$D_{H_L} = \{(h_L, \rho(h_L)) | h_L \in H_L\} \tag{12}$$

where $\rho(h_L) \in [0,1] \cup \{null\}$ and $\rho(h_L)$ is the symbolic proportion of $h_L$ if $\rho(h_L) \neq \{null\}$.

In an HLD $D_{H_L}$, preference from a decision maker is a set of HFLTSs $h_L$ with certain possibility $\rho(h_L)$, and the element $h_L$ is not used if $\rho(h_L) = \{null\}$. If $\sum_{h_L \in H_L} \rho(h_L) = 1$, $D_{H_L}$ is complete; if $\sum_{h_L \in H_L} \rho(h_L) < 1$, $D_{H_L}$ is with incompleteness. Zhang et al. [109] proposed a normalization method for the case of $\sum_{h_L \in H_L} \rho(h_L) > 1$.

## 2.6 Taxonomy of distributed linguistic representations

We present the taxonomy of the existing distributed linguistic representations. Based on the concepts of LD and its variants, their comparison results are listed in Table 1.

Table 1. The analysis of distributed linguistic representations

| Linguistic expression | Mathematical format | Symbolic proportion information | Distribution over linguistic information | Incomplete information considered |
|---|---|---|---|---|
| LD [107] | $\{(l_i, \rho(l_i)) | l_i \in L\}$ | $\sum_{l_i \in L} \rho(l_i) = 1$ | Terms in $L$ | No |
| INLD [17] | $\{(l_i, \rho(l_i)) | l_i \in L\}$ | $[\rho^L(l_i), \rho^U(l_i)]$ | Terms in $L$ | No |

| | | | | |
|---|---|---|---|---|
| PDHFLTS [90] | $\{(l_i, p_i) \mid l_i \in h_L\}$ | $\sum_{l_i \in h_L} p_i = 1$ | Terms in an HFLTS | No |
| PHFLTS [9] | $\{(l_i, p_i) \mid l_i \in L\}$ | $\sum_{l_i \in L} p_i = 1$ | Terms in $L$ | Yes |
| PLTS [66] | $\{l_i(p_i) \mid l_i \in L\}$ | $\sum_{i=0}^{n(L(p))} p_i \leq 1$ | Terms in $L$ | Yes |
| PLD [26] | $\{(\gamma_i l_i, ..., \gamma_{i+r} l_{i+r}, \varepsilon) \mid l_i \in L\}$ | $\sum_{j=i}^{i+r} \gamma_i \leq 1$ | Terms in $L$ | Yes |
| HLD [109] | $\{(h_L, \rho(h_L))\}$ | $\rho(h_L) \in [0,1]$ | HFLTS of $L$ | Yes |
| ILD [105] | $\{(l_i, \rho(l_i)), \eta \mid l_i \in L\}$ | $\sum_{l_i \in L} \rho(l_i) + \eta = 1$ | Terms in $L$ | Yes |
| FLE [86] | $\{(s_L, \rho(s_L))\}$ | $\rho(s_L) \in [0,1]$ | Subset of $L$ | Yes |

From the above classifications, we can figure out the following characteristics:

(1) FLE is the generalization of almost all distributed linguistic representations reviewed.

(2) LD, PDHFLTS, PHFLTS, PLTS, PLD, and ILD, are special HLDs.

(3) PLTS is mathematically consistent with ILD; and PLD shares high similarity with PLTS and ILD, but in PLD symbolic proportions are distributed over successive linguistic terms. Because this difference is minor, in this review we think that PLD is equivalent to PLTS and ILD (approximately).

(4) PHFLTS is mathematically consistent with LD.

(5) PDHFLTS is the special case of LD.

(6) INLD is a generalization of LD.

The relationships among different distributed linguistic representations can be described as Fig. 2.

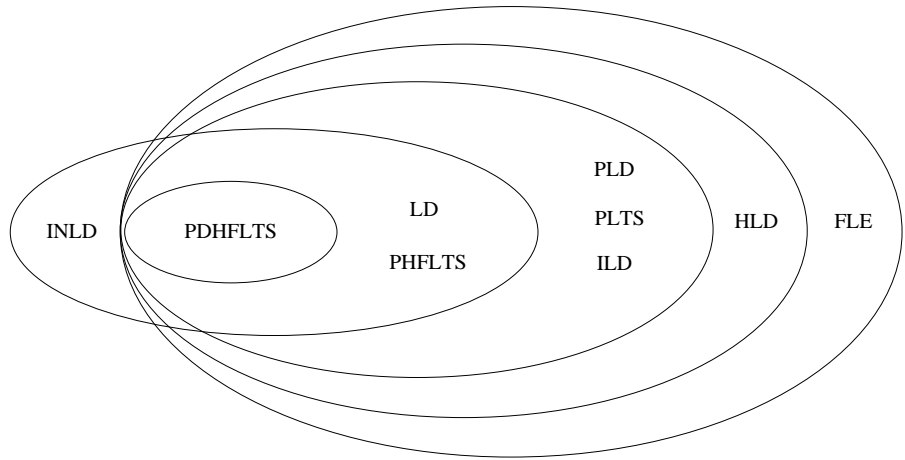

Fig.2 The taxonomy of distributed linguistic representations

**Remark 2.** To our knowledge, LD is the first attempt in the linguistic decision making community to model the distributed linguistic representations. Although the follow-up notions and concepts are mathematically consistent with LD in some sense, they are designed with different purposes:

(1) The notions of PDHFLTS, PHFLTS, PLTS, HLD and FLE are designed for

extending HFLTS, while PLD, ILD and INLD are designed for extending proportional linguistic 2-tuples.

(2) PDHFLTS is designed to the modelling of individual linguistic expression, and PHFLTS is established for group linguistic representations as it involves the fusion of individuals' linguistic preference information.

## 3. Key elements and applications of distributed linguistic representations in decision making

In this section, we show the key elements and applications of distributed linguistic representations in decision making. Specifically, distance measurements and aggregation methods under various distributed linguistic representations are summarized in Section 3.1 and 3.2 respectively. Based on the use of distributed linguistic representations, several distributed linguistic preference relations are developed, which are introduced in Section 3.3. Following this, multiple attribute decision making (MADM) models with distributed linguistic representations are presented in Section 3.4. Finally, some real-life applications are reviewed in Section 3.5.

### 3.1. Distributed linguistic distance measurements

In this section, we review several widely used distance measurements in distributed linguistic representations. To start with, in Table 2 we provide a summary of these distance measurements (refer to column 4), which are the bases to support decision making.

Table 2: A summary of distributed linguistic distance measurements and aggregation

| Distributed linguistic representations | References | Features of linguistic terms | Distance measurement | Aggregation method | Aggregation result |
|---|---|---|---|---|---|
| LD | [107] | Unified, balanced | Symbolic proportions based extended Manhattan distance | LDWA LDOWA | LD |
| | [119] | Multi-granular, balanced | Expectation based extended Manhattan distance | LDWA | LD |
| | [88] | Unified, balanced | Not discussed | Maximum support degree model with accuracy | HFLTS |
| INLD | [17] | Multi-granular, unbalanced | Not discussed | INLDWA INLDOWA | INLD |
| ILD | [105] | Unified, balanced | Extended Manhattan distance considering symbolic | Consensus-oriented aggregation model | ILD |

| | | | proportions and uncertainty degree | | |
|---|---|---|---|---|---|
| PLTS | [66] | Unified, balanced | Extended Euclidean distance | PLTSWA PLTS weighted geometric operator | PLTS |
| | [28] | Unified, unbalanced | Weighted distance, Weighted Hausdorff distance, Hybrid weighted Hamming distance | PLTSWA based on Archimedean copula | PLTS |
| | [49] | Unified, balanced | Extended Manhattan distance considering probability | PLTSWA PLTSOWA | PLTS |
| HLD | [109] | Unified, balanced | Not discussed | HLDWA HLDOWA | HLD |
| PDHFLTS | [90] | Unified, balanced | Expectation based extended Manhattan distance | PDHFLWA PDHFLOWA | PDHFLTS |
| FLE | [86] | Unified, balanced | Symbolic proportions based extended Manhattan distance | Aggregation model with accuracy and minimum preference-loss | LD |

Next, the main distance measurements of LD, PDHFLTS, PLTS, ILD, and FLE are reviewed.

**(1) Distance measurements of LD and PDHFLTS**

Let $D_L^r = \{(l_i, \rho^r(l_i)) | i=0,1,...,g\}$ and $D_L^s = \{(l_i, \rho^s(l_i)) | i=0,1,...,g\}$ be two LDs. Several methods for distance measurements in LDs were reported (see [33], [97], [107], [119]).

The distance between $D_L^r$ and $D_L^s$ proposed by Zhang et al. [107] is provided as follows:

$$d(D_L^r, D_L^s) = \frac{1}{2} \sum_{i=0}^{g} | \rho^r(l_i) - \rho^s(l_i) |. \tag{13}$$

Zhang et al. [119] pointed out that the above distance measurement defined by Eq. (13) just calculates the deviation between symbolic proportions and ignores the linguistic terms. To overcome this drawback, Zhang et al. [119] proposed another distance measurement as follows:

$$d(D_L^r, D_L^s) = \frac{1}{g} | \sum_{i=0}^{g} (\rho^r(l_i) - \rho^s(l_i)) \cdot NS(l_i) | \tag{14}$$

where $NS(\cdot)$ is the numerical scale [18], and $NS(l_i)$ is set to be *i*.

Ju et al. [33] proposed the following way to calculate the distance between LDs

$D_L^r$ and $D_L^s$:

$$d(D_L^r, D_L^s) = \sqrt{\sum_{i=0}^{g}(\rho^r(l_i) \cdot NS(l_i) - \rho^s(l_i) \cdot NS(l_i))^2}. \quad (15)$$

Similar to Eq. (14), $NS(l_i)$ in Eq. (15) is set to be $i$.

Yao [97] used the following way to compute the distance between LDs $D_L^r$ and $D_L^s$:

$$d(D_L^r, D_L^s) = \frac{1}{g}\sum_{t=0}^{g}|\sum_{i=0}^{t}\rho^r(l_i) - \sum_{i=0}^{t}\rho^s(l_i)|. \quad (16)$$

Wu and Xu [90] discussed the distance measurement of PDHFLTSs. Let $D_{h_L^r}^{P,r} = \{(l_i, p_i^r) | l_i \in L\}$ and $D_{h_L^s}^{P,s} = \{(l_i, p_i^s) | l_i \in L\}$ be two PDHFLTSs. The distance between $D_{h_L^r}^{P,r}$ and $D_{h_L^s}^{P,s}$ is defined by:

$$d(D_{h_L^s}^{P,s}, D_{h_L^s}^{P,s}) = \frac{1}{g}|\sum_{i=0}^{g} p_i^r \cdot NS(l_i) - \sum_{i=0}^{g} p_i^s \cdot NS(l_i)| \quad (17)$$

where $NS(l_i) = i$, and Eq. (17) is equivalent to Eq. (14).

**(2) Distance measurements of PLTS and ILD**

Let $L^r(p) = \{(l_i^r(p_i^r)) | l_i^r \in L, p_i^r \geq 0, i = 1, ..., n(L^r(p))\}$ and $L^s(p) = \{(l_i^s(p_i^s)) | l_i^s \in L, p_i^s \geq 0, i = 1, ..., n(L^s(p))\}$ be two PLTSs, and $n(L^r(p)) = n(L^s(p))$. Pang et al. [66] proposed the deviation degree between $L^r(p)$ and $L^s(p)$:

$$d(L^r(p), L^s(p)) = \sqrt{\sum_{i=1}^{n(L^r(p))}(p_i^r \cdot su_i^r(l_i^r) - p_i^s \cdot su_i^s(l_i^s))^2 / n(L^r(p))} \quad (18)$$

where $su_i^r$ and $su_i^s$ are the subscripts of linguistic terms $l_i^r$ and $l_i^s$ respectively.

Zhang et al. [105] analyzed the distance measurement of ILDs. Let $D_L^{I,r} = \{(l_i, \rho^r(l_i)), \eta^r | i = 0, 1, ..., g\}$ and $D_L^{I,s} = \{(l_i, \rho^s(l_i)), \eta^s | i = 0, 1, ..., g\}$ be two ILDs. Then, the distance measurement between ILDs $D_L^{I,r}$ and $D_L^{I,s}$ is given by:

$$d(D_L^{I,r}, D_L^{I,s}) = \frac{1}{2}(\sum_{i=0}^{g}|\rho^r(l_i) - \rho^s(l_i)| + \delta|\eta^r - \eta^s|) \quad (19)$$

where $\delta \in (0,1]$ is a predefined distance parameter, and represents the importance degree of incomplete information in the distance measurement.

**(3) Distance measurements of FLE**

Wu et al. [86] proposed the distance measurement for FLEs. Let $f_L = \{(s_L, \rho(s_L)) | s_L \in S_L\}$ be an FLE, and $D_{h_L}^P = \{(l_i, \rho(l_i)) | i = 0, 1, ..., g\}$ be a PDHFLTS of $L$. The distance between $f_L$ and $D_{h_L}^P$ is defined by:

$$d(f_L, D_{h_L}^P) = \sum_{s_L \in S_L} |\sum_{l_i \in s_L} \rho(l_i) - \rho(s_L)| \quad (20)$$

where $\rho(s_L) \in [0,1]$ is the symbolic proportion of the subset $s_L$ in $f_L$, and $\sum_{l_i \in s_L} \rho(l_i)$ is the sum of symbolic proportions of all the simple terms in $s_L$ in $D_{h_L}^P$.

## 3.2. Aggregation approaches of distributed linguistic representations

In this section, we review the commonly used aggregation methods for distributed linguistic representations, which are summarized in Table 2 (columns 5 and 6).

**(1) Aggregation approaches in LD, INLD**

Let $\{D_L^1, D_L^2, ..., D_L^m\}$ be a set of LDs of $L$, where $D_L^k = \{(l_i, \rho^k(l_i)) | i = 0, 1, ..., g\}$, and $\lambda = (\lambda_1, \lambda_2, ..., \lambda_m)^T$ be an associated weighting vector that satisfies $\lambda_k > 0$ and $\sum_{k=1}^m \lambda_k = 1$. Zhang et al. [107] proposed the weighted averaging operator for LDs by:

$$LDWA_\lambda(D_L^1, D_L^2, ..., D_L^m) = \{(l_i, \rho^c(l_i)) | i = 0, 1, ..., g\} \tag{21}$$

where $\rho^c(l_i) = \sum_{k=1}^m \lambda_k \rho^k(l_i)$, $i = 0, 1, ..., g$.

The ordered weighted averaging operator of LDs is computed by [107]:

$$LDOWA_\lambda(D_L^1, D_L^2, ..., D_L^m) = \{(l_i, \rho^c(l_i)) | i = 0, 1, ..., g\} \tag{22}$$

where $\rho^c(l_i) = \sum_{k=1}^m \lambda_k \rho^{\sigma(k)}(l_i)$ and $\{\sigma(1), \sigma(2), ..., \sigma(m)\}$ is a permutation of $\{1, 2, ..., m\}$ such that $D_L^{\sigma(k-1)} \geq D_L^{\sigma(k)}$ for $k = 2, 3, ..., m$.

Liang et al. [44] defined other operators for the aggregation of LDs. Let $\{D_L^1, D_L^2, ..., D_L^m\}$ be as above. The LD power average (LDPA) operator is defined by:

$$LDPA(D_L^1, D_L^2, ..., D_L^m) = \{(l_i, \rho^c(l_i)) | i = 0, 1, ..., g\} \tag{23}$$

where

$$\rho^c(l_i) = \sum_{k=1}^m \rho^k(l_i) \frac{(1 + T(D_L^k))}{\sum_{k=1}^m (1 + T(D_L^k))} \tag{24}$$

and $T(D_L^k) = \sum_{r=1, r \neq k}^m Sup(D_L^k, D_L^r)$, $Sup(D_L^k, D_L^r) = 1 - d(D_L^k, D_L^r)$ is the support degree of the elements $D_L^k$ and $D_L^r$.

Based on the LDPA operator, the LD weighted power average (LDWPA) operator is defined by [44]:

$$LDWPA_\lambda(D_L^1, D_L^2, ..., D_L^m) = \{(l_i, \rho^c(l_i)) | i = 0, 1, ..., g\} \tag{25}$$

where

$$\rho^c(l_i) = \sum_{k=1}^m \rho^k(l_i) \frac{\lambda_k(1 + T(D_L^k))}{\sum_{k=1}^m \lambda_k(1 + T(D_L^k))}. \tag{26}$$

Dong et al. [17] proposed the aggregation methods for INLDs. Let $\{\tilde{D}_L^1, \tilde{D}_L^2, ..., \tilde{D}_L^m\}$ be a set of INLDs of $L$, where $\tilde{D}_L^k = \{(l_i, \rho^k(l_i)) | i = 0, 1, ..., g\}$. The weighted average operator of INLDs is computed by:

$$INLDWA_\lambda(\tilde{D}_L^1, \tilde{D}_L^2, ..., \tilde{D}_L^m) = \{(l_i, [\underline{\rho}^c(l_i), \overline{\rho}^c(l_i)]) | i = 0, 1, ..., g\} \tag{27}$$

where $\underline{\rho}^c(l_i) = \sum_{k=1}^m \lambda_k \cdot \underline{\rho}^k(l_i)$, $\overline{\rho}^c(l_i) = \sum_{k=1}^m \lambda_k \cdot \overline{\rho}^k(l_i)$. $\lambda_k$ is the weighting vector of

$\tilde{D}_L^k$ satisfying $\sum_{k=1}^m \lambda_k = 1$.

The ordered weighted average operator of INLDs is computed by [17]:

$$INLDOWA_\lambda(\tilde{D}_L^1, \tilde{D}_L^2, ..., \tilde{D}_L^m) = \{(l_i, [\underline{\rho}^c(l_i), \bar{\rho}^c(l_i)]) | i = 0,1,...,g\} \quad (28)$$

where $\underline{\rho}^c(l_i) = \sum_{k=1}^m \lambda_k \cdot \underline{\rho}^{\sigma(k)}(l_i)$, $\bar{\rho}^c(l_i) = \sum_{k=1}^m \lambda_k \cdot \bar{\rho}^{\sigma(k)}(l_i)$. $\{\sigma(1), \sigma(2), ..., \sigma(m)\}$ is a permutation of $\{1, 2, ..., m\}$ such that $\tilde{D}_L^{\sigma(k-1)} \geq \tilde{D}_L^{\sigma(k)}$ for $k = 2,3,...,m$.

Based on the use of LDs and HFLTSs, Wu et al. [88] proposed the maximum support degree model (MSDM), aiming at maximizing the support degree as well as guarantying the accuracy of group opinion, which is presented below:

$$\begin{cases} \max \phi(h_L^c) \\ s.t. \begin{cases} \phi(h_L^c) = \sum_{l_i \in h_L^c} \phi(l_i) \\ \phi(l_i) = \sum_{k=1}^m \lambda_k \rho^k(l_i) \\ n(h_L^c) \leq T \\ h_L^c \in H_L \end{cases} \end{cases} \quad (29)$$

where $\phi(h_L^c)$ is the support degree of the collective opinion $h_L^c$, $\phi(l_i)$ is the support degree of the linguistic term $l_i$ in $D_L^k$. The constraint $n(h_L^c) \leq T$ guarantees that the number of linguistic terms in $h_L^c$ is less than a preset value $T$, and constraint $h_L^c \in H_L$ guarantees that $h_L^c$ is an HFLTS.

**(2) Aggregation approaches in PLD, PLTS, and ILD**

Let

$$\Re = \begin{Bmatrix} (\gamma_{i_1}^1 l_{i_1}, \gamma_{i_1+1}^1 l_{i_1+1}, ..., \gamma_{i_1+r_1}^1 l_{i_1+r_1}, \varepsilon_1 | l_{i_t} \in L), \\ \cdots, \\ (\gamma_{i_m}^m l_{i_m}, \gamma_{i_m+1}^m l_{i_m+1}, ..., \gamma_{i_m+r_m}^m l_{i_m+r_m}, \varepsilon_1 | l_{i_t} \in L) \end{Bmatrix}$$

be a set of PLDs, and $D_L^{p,k} = \{(\gamma_{i_k}^k l_{i_k}, \gamma_{i_k+1}^k l_{i_k+1}, ..., \gamma_{i_k+r_k}^k l_{i_k+r_k}, \varepsilon_k) | l_{i_k} \in L\}$ $(k=1,...,m)$. Then, the weighted average of PLDs in $\Re$, which is also a PLD denoted by $(\gamma_i^c l_i, \gamma_{i+1}^c l_{i+1}, ..., \gamma_{i+r}^c l_{i+r}, \varepsilon_c | l_i,...l_{i+r} \in L)$, is defined by Guo et al. [26]:

$$\begin{cases} \gamma_j^c l_j = (\sum_{k=1}^m \lambda_k \gamma_j^k) l_j, j \in \{i, i+1, ..., i+r\} \\ \varepsilon_c = \sum_{k=1}^m \lambda_k \varepsilon_k \end{cases} \quad (30)$$

where $i$ is the minimum index of starting labels and $i+r$ is the maximum index of ending labels of PLDs in $\Re$ respectively, i.e., $i = \min\{i_1,...,i_m\}$ and $i+r = \max\{i_1+r_1,...,i_m+r_m\}$.

In addition, the weighted average operator for PLDs defined above can also be extended for the case where the weights are expressed by means of uncertain linguistic weights instead of numerical values [26].

Different aggregation methods have been proposed for dealing with PLTSs. Pang et al. [66] and Liu et al. [49] defined the PLTS weighted averaging (PLTSWA) operator respectively, and Liu et al. [49] also proposed ordered weighted averaging operator of PLTSs (PLTSOWA). In addition, Han et al. [28] presented the concepts of Archimedean copula weighted probabilistic unbalanced linguistic arithmetic average aggregation operator and Archimedean copula weighted probabilistic unbalanced linguistic geometric average aggregation operator.

Let $\{D_L^{I,1}, D_L^{I,2}, ..., D_L^{I,m}\}$ be a set of ILDs, where $D_L^{I,k} = \{(l_i, \rho^k(l_i)), \eta^k \mid i = 0, 1, ..., g\}$. Let $D_L^{I,c} = \{(l_i, \rho^c(l_i)), \eta^c \mid i = 0, 1, ..., g\}$ be the collective ILD, Zhang et al. [105] proposed the following model to obtain $D_L^{I,c}$:

$$\min \frac{1}{2mn} \sum_{k=1}^{m} \sum_{q=1}^{n} (\sum_{i=0}^{g} |\rho_q^k(l_i) - \rho_q^c(l_i)| + \delta |\eta_q^k - \eta_q^c|) \quad (31)$$

where $\rho_q^k(l_i)$ is the symbolic proportion information of $l_i$ provided by individual $i$ for alternative $q$. $\delta$ represents the importance degree of incomplete information in the distance measurement of $D_L^{I,k}$ and $D_L^{I,c}$.

Meanwhile, since $D_L^{I,c}$ is an ILD, we have that $\sum_{i=0}^{g} \rho^c(l_i) + \eta^c = 1$, $\rho^c(l_i) \in [0,1]$ and $\eta^c \in [0,1]$. In order to get relatively precise collective opinion, the incomplete degree of $D_L^{I,c}$ is set to be less than the average incompleteness of all decision makers' opinions, i.e., $\eta^c \in [0, \frac{1}{m} \sum_{k=1}^{m} \eta^k]$.

**(3) Aggregation approaches in PDHFLTS, PHFLTS, HLD, and FLE**

In [90], Wu and Xu defined the hesitant fuzzy linguistic weighted average (HFLWA) operator and hesitant fuzzy linguistic ordered weighted average (HFLOWA) operator for the aggregation of PDHFLTSs.

Chen et al. [9] defined the proportional HFLWA (PHFLWA) operator and proportional HFLOWA (PHFLOWA) operator for the aggregation of PHFLTSs. Let $\{P_{H_L}^1, P_{H_L}^2, ..., P_{H_L}^m\}$ be a set of PHFLTSs. The PHFLWA operator is defined by [9]:

$$\begin{aligned} PHFLWA(P_{H_L}^1, P_{H_L}^2, ..., P_{H_L}^m) &= TG^m(\lambda_k, P_{H_L}^k \mid k = 1, 2, ..., m) \\ &= T((\lambda_1, P_{H_L}^1) \otimes TG^{m-1}(\lambda_k, P_{H_L}^k \mid k = 2, ..., m)) \end{aligned} \quad (32)$$

The PHFLOWA operator is defined as follows [9]:

$$\begin{aligned} PHFLOWA(P_{H_L}^1, P_{H_L}^2, ..., P_{H_L}^m) &= TG^m(\lambda_k, P_{H_L}^{\sigma(k)} \mid k = 1, 2, ..., m) \\ &= T((\lambda_1, P_{H_L}^1) \otimes TG^{m-1}(\lambda_k, P_{H_L}^{\sigma(k)} \mid k = 2, ..., m)) \end{aligned} \quad (33)$$

where $\{\sigma(1), \sigma(2), ..., \sigma(m)\}$ is a permutation of $\{1, 2, ..., m\}$ such that $P_{H_L}^{\sigma(r)} \succ P_{H_L}^{\sigma(s)}$. $T$ is a triangular norm [3], [71], and $G(\cdot)$ is a proportional convex combination function [9].

Zhang et al. [109] proposed the weighted average operator and ordered weighted average operator for the aggregation of HLDs. Let $D_{H_L}^1 = \{(h_L^1, \rho^1(h_L^1)) \mid h_L^1 \in H_L\}$ and $D_{H_L}^2 = \{(h_L^2, \rho^2(h_L^2)) \mid h_L^2 \in H_L\}$ be two HLDs. Let $\lambda_1$ and $\lambda_2$ be the corresponding weights, where $\lambda_1, \lambda_2 \geq 0$ and $\lambda_1 + \lambda_2 = 1$. Then the weighted union of $D_{H_L}^1$ and $D_{H_L}^2$ is defined by [109]:

$$U(D_{H_L}^1, D_{H_L}^2) = U(\lambda_1 D_{H_L}^1, \lambda_2 D_{H_L}^2) \tag{34}$$

where $U(\lambda_1 D_{H_L}^1, \lambda_2 D_{H_L}^2) = \{(h_L^c, \rho^c(h_L^c)) \mid h_L^c \in H_L\}$, with

$$(h_L^c, \rho^c(h_L^c)) = \begin{cases} (h_L^1, \lambda_1 \rho^1(h_L^1) + \lambda_2 \rho^2(h_L^2)), & \text{if } \rho^1(h_L^1), \rho^2(h_L^2) \neq null, h_L^1 = h_L^2 \\ (h_L^1, \lambda_1 \rho^1(h_L^1)) \cup (h_L^2, \lambda_2 \rho^2(h_L^2)), & \text{if } \rho^1(h_L^1), \rho^2(h_L^2) \neq null, h_L^1 \neq h_L^2 \\ (h_L^1, \lambda_1 \rho^1(h_L^1)), & \text{if } \rho^1(h_L^1) \neq null, \rho^2(h_L^2) = null \\ (h_L^2, \lambda_2 \rho^2(h_L^2)), & \text{if } \rho^2(h_L^2) \neq null, \rho^1(h_L^1) = null \end{cases} \tag{35}$$

Let $\{D_{H_L}^1, D_{H_L}^2, ..., D_{H_L}^m\}$ be a set of HLDs, where $D_{H_L}^k = \{(h_L^k, \rho^k(h_L^k)) \mid h_L^k \in H_L\}$ ($k = 1, 2, ..., m$). The weighted average operator for $\{D_{H_L}^1, D_{H_L}^2, ..., D_{H_L}^m\}$ is defined by [109]:

$$HLDWA_\lambda(D_{H_L}^1, D_{H_L}^2, ..., D_{H_L}^m) = U(\lambda_1 D_{H_L}^1, U(\lambda_2 D_{H_L}^2, ..., \lambda_m D_{H_L}^m)). \tag{36}$$

The ordered weighted average operator of $\{D_{H_L}^1, D_{H_L}^2, ..., D_{H_L}^m\}$ is defined by [109]:

$$HLDOWA_\lambda(D_{H_L}^1, D_{H_L}^2, ..., D_{H_L}^m) = U(\lambda_1 D_{H_L}^{\sigma(1)}, U(\lambda_2 D_{H_L}^{\sigma(2)}, ..., \lambda_m D_{H_L}^{\sigma(m)})) \tag{37}$$

where $(\sigma(1), \sigma(2), ..., \sigma(m))$ is a permutation of $\{1, 2, ..., m\}$ such that $D_{H_L}^{\sigma(k-1)} > D_{H_L}^{\sigma(k)}$.

Wu et al. [86] proposed an optimization approach for the aggregation of FLEs. Let $f_L^k = \{(s_L^k, \rho(s_L^k)) \mid s_L^k \in S_L^k\}$ be the FLE preference of individual $k$ ($k = 1, 2, ..., m$), and $D_L^{P,c}$ be the collective PDHFLTS. $h_L^c$ is the HFLTS associated with $D_L^{P,c}$ [86]:

$$\begin{aligned} & \min \sum_{k=1}^m \lambda_k d(f_L^k, D_L^{P,c}) \\ & s.t. \begin{cases} D_L^{P,c} \in PD_L \\ n(h_L^c) \leq T \end{cases} \end{aligned} \tag{38}$$

where $\lambda_k$ is the weighting vector of individual $k$, and $n(h_L^c)$ is the number of elements in $h_L^c$.

### 3.3. Distributed linguistic preference relation

Based on the use of distributed linguistic representations in the pairwise comparison method, several distributed linguistic preference relations including linguistic distribution preference relation (LDPR), probabilistic linguistic preference relation (PLPR), and flexible linguistic preference relation (FLPR) have been reported, which are summarized in Table 3.

Table 3: A summary of distributed linguistic preference relations

| Distributed linguistic preference relations | Distributed linguistic representations | References | Complete vs Incomplete |
|---|---|---|---|
| LDPR | LD | [107] | Complete |

|      |      | [74]  | Incomplete |
|------|------|-------|------------|
|      |      | [92]  | Complete   |
|      |      | [75]  | Complete   |
|      |      | [76]  | Complete   |
|      |      | [73]  | Complete   |
|      |      | [49]  | Complete   |
| PLPR | PLTS | [115] | Complete   |
|      |      | [21]  | Incomplete |
|      |      | [52]  | Complete   |
|      |      | [116] | Complete   |
| FLPR | FLE  | [87]  | Complete   |

The details of these distributed linguistic distribution preference relations are introduced in the rest of this section.

For the ease of illustration, all distributed linguistic preference relations involved in this review are denoted as $A=(a_{rj})_{n\times n}$.

**Definition 13 (LDPR) [107]**: An LDPR on the set of alternatives $X=\{x_1,x_2,...x_n\}$ is represented by a matrix $A=(a_{rj})_{n\times n}$, where $a_{rj}=\{(l_i,\rho_{rj}(l_i))\,|\,i=0,1,...,g\}$ is an LD of $L$, and represents the preference degree of alternative $x_r$ over $x_j$. $A=(a_{rj})_{n\times n}$ is reciprocal if $Neg(a_{rj})=a_{jr}$, for $r,j=1,2,...,n$.

Particularly, if $\sum_{i=0}^{g}\rho_{rj}(l_i)=1$ for all $r,j\in\{1,2,...,n\}$, then $A$ is called LDPR with complete symbolic proportions; otherwise, $A$ is called LDPR with incomplete symbolic proportions [74].

In many cases, words mean different things to different decision makers. In this case, the LD preference relations are called PIS-based LD preference relations [75], [76], [92].

**Definition 14 (PLPR) [115], [116]**: A PLPR on the set of alternatives $X=\{x_1,x_2,...x_n\}$ is represented by a matrix $A=(a_{rj})_{n\times n}$, where $a_{rj}=\{l_{i,rj}(p_{i,rj})\,|\,l_{i,rj}\in L, p_{i,rj}\geq 0, i=1,...,n(a_{rj})\}$ is a PLTS, and represents the preference degree of alternative $x_r$ over $x_j$, $r,j\in\{1,2,...,n\}$.

Moreover, incomplete PLPR is developed in [21].

**Definition 15 (FLPR) [87]**: An FLPR on the set of alternatives $X=\{x_1,x_2,...x_n\}$ is represented by a matrix $A=(a_{rj})_{n\times n}$, where $a_{rj}=\{(s_L,\rho_{rj}(s_L))\,|\,s_L\in S_{L,rj}\}$ is an FLE, $Neg(a_{rj})=a_{jr}$, and $a_{rj}$ represents the preference degree of alternative $x_r$ over $x_j$. $S_{L,rj}$ is the set whose elements $s_L$ are the subsets of $L$, and the decision maker uses the elements of $S_{L,rj}$ to express his/her preference of alternative $x_r$ over $x_j$.

The transformations between different distributed linguistic preference relations are investigated by several researchers [87], [112], [114].

### 3.4. Distributed linguistic MADM

The distributed linguistic representations have been utilized in MADM to model the uncertain assessments of decision makers, and several distributed linguistic MADM approaches are developed accordingly, which are summarized in Table 4.

Table 4: A summary of distributed linguistic MADM

| Distributed linguistic MADM | References | Decision context | Features of linguistic terms |
|---|---|---|---|
| MADM with LD | [97] | Group | Unified, balanced |
| | [84] | Group | Unified, balanced |
| | [88] | Group | Unified, balanced |
| | [119] | Large-scale group | Multi-granular, balanced |
| | [120] | Large-scale group | Multi-granular, unbalanced |
| | [46] | Group | Multi-granular, balanced |
| | [99] | Large-scale group | Multi-granular, unbalanced |
| MADM with INLD | [17] | Group | Multi-granular, unbalanced |
| MADM with PDHFLTS | [90] | Group | Unified, balanced |
| | [91] | Group | Unified, balanced |
| | [8] | Group | PIS |
| MADM with PHFLTS | [9] | Group | Unified, balanced |
| MADM with PLD | [26] | Individual | Balanced |
| MADM with PLTS | [28] | Group | Unified, unbalanced |
| | [51] | Group | Unified, balanced |
| | [50] | Group | Unified, balanced |
| | [66] | Group | Unified, balanced |
| | [77] | Individual | Balanced |
| | [83] | Group | Unified, balanced |
| | [81] | Group | Multi-granular, balanced |
| MADM with ILD | [105] | Group | Unified, balanced |
| | [113] | Group | Unified, balanced |
| MADM with HLD | [109] | Group | Unified, balanced |
| MADM with FLE | [86] | Group | Unified, balanced |

**(1) MADM approaches with LD and INLD**

The LD has been applied in MADM to represent uncertain assessments of decision makers. Yao [97] proposed a consensus reaching model for multiple attribute group decision making (MAGDM) with assessments represented by means of LDs, in which a feedback mechanism is devised by combining an identification rule and an optimization-based model. Wu et al. [84] presented a minimum adjustment cost feedback mechanism based consensus model for MAGDM under social network, in which the assessments information and the trust relationships among decision makers are both represented by LDs. Based on the use of LDs and HFLTSs, Wu et al. [88] proposed the MSDM to address linguistic MAGDM problem, which aims at maximizing the support degree of group opinion as well as guarantying the accuracy of group opinion. Zhang et al. [119] designed an approach to manage multi-granular LDs in large-scale MAGDM, in which a linguistic computational model is developed based on the extended linguistic hierarchies model and the transformation formulas between a linguistic 2-tuple and an LD. Recently, Zhang et al. [120] developed a large-scale MAGDM model with multi-granular unbalanced hesitant fuzzy linguistic information. In their model, all unbalanced hesitant fuzzy linguistic information is transformed into LDs defined on a balanced linguistic term set. Yu et al. [99] proposed a method to deal with large-scale GDM problems with multi-granular unbalanced linguistic information, in which the initial multi-granular unbalanced linguistic information of decision makers is represented by unbalanced LDs and the classical TODIM method is extended to derive a raking of alternatives. Liang et al. [46] developed a consensus-based analysis model for MAGDM with multi-granular LD preferences.

Dong et al. [17] proposed an MAGDM with INLD under multi-granular unbalanced linguistic contexts. First, a basic linguistic term set is selected to normalize the individual unbalanced INLDs matrices. Then, the collective unbalanced INLDs matrix is obtained by aggregating the normalized individual unbalanced INLDs matrices. Following this, the ranking of alternatives is obtained from the collective unbalanced INLDs matrix.

**(2) MADM approaches with PLD, PLTS, and ILD**

In real-world MADM, due to the limitation of knowledge, problem complexity and time pressure, the assessment information given by decision makers may not be complete. Various distributed linguistic MADM methods to handle incomplete assessment information have been proposed.

Guo et al. [26] proposed a PLD based model for MADM under linguistic uncertainty, which is based on the nature of symbolic linguistic model combined with distributed assessments. Moreover, this model is also able to deal with ILDs so that it allows evaluators to avoid the dilemma of having to supply complete assessments when not available.

Han et al. [28] proposed a new computational model based on Archimedean copula for unbalanced PLTS and developed an MAGDM based on it. Liu et al. [51] presented the bidirectional projection method for probabilistic linguistic MAGDM based on power average operator. Liu and Li [50] developed an extended MULTIMOORA method for probabilistic linguistic MAGDM based on prospect theory. Pang et al. [66] developed an extended TOPSIS method and an aggregation-based method respectively for MAGDM with probabilistic linguistic information. Tian et al. [77]

presented a probabilistic linguistic MADM based on evidential reasoning and combined ranking methods considering decision-makers' psychological preferences. Wang et al. [83] proposed a distance-based MAGDM approach with PLTSs. Wang [81] proposed a generalized distance measurements method between two PLTs with multi-granular linguistic information, and applied them to deal with multi-granular MAGDM problems. Zhou et al. [123] proposed particle swarm optimization method for trust relationship based social network MAGDM under a probabilistic linguistic environment.

Zhang et al. [105] developed a consensus-oriented aggregation model for MAGDM with ILDs, which can obtain a collective opinion with maximum consensus, and further developed a minimum-cost consensus model with variable unit consensus cost. Zhang et al. [113] developed a deviation minimum-based optimization model to manage ILDs in MAGDM by minimizing the opinion deviation among decision makers, and proposed a consensus-reaching model with bounded confidences based feedback adjustment mechanism to assist decision makers to gain a consensus.

**(3) MADM approaches with PDHFLTS, PHFLTSs, HLD, and FLE**

HFLTS based distributed linguistic MADM approaches have been proposed and utilized to tackle a variety of decision problems.

Wu and Xu [90] proposed operation laws and aggregation operators and for PDHFLTS and built a framework to deal with both consensus and selection processes for MAGDM problems with PDHFLTSs. Wu et al. [91] examined an MAGDM problem in which the linguistic information was represented by PDHFLTS, and developed two approaches based on VIKOR and TOPSIS to find a compromise solution.

Chen et al. [9] developed a proportional hesitant fuzzy linguistic MAGDM model, in which a probability theory-based outranking method for PHFLTSs was proposed and two fundamental aggregation operators for PHFLTSs were provided.

Zhang et al. [109] proposed an approach for MAGDM with HLD, in which the transformation between HLDs and LDs and the basic comparison and aggregation operations to perform on HLDs are developed.

Wu et al. [86] proposed MAGDM with FLE. In the proposed model, an FLE aggregation process with accurate constraints is developed to improve the quality (i.e., accuracy) of the collective result as well as guarantee the principle of minimum preference-loss through a mixed 0-1 linear programming model. Meanwhile, the consensus rules with minimum preference-loss are designed to support the consensus reaching process in the MAGDM with FLE.

**3.5. Some real-life applications**

In this section, some real-life applications of the distributed linguistic representations are introduced. In Table 5, we provide a summary of them.

Table 5: A summary of distributed linguistic representations in real-life applications

| Applications | References | Distributed linguistic representations |
|---|---|---|
| Failure mode and effect analysis | [32] | LD |
| | [34] | LD, PLTS |
| | [64] | Multi-granular LD |
| | [110] | LD |
| | [111] | LD with PIS |
| | [113] | ILD |
| Hotel selection | [43] | LD |
| | [62] | LD |
| | [67] | PLTS |
| | [98] | LD |
| | [106] | LD |
| Water security sustainability evaluation | [63] | LD |
| Sustainability of constructed wetlands evaluating | [52] | PLTS |
| Sustainable third-party reverse logistics provider selection | [53] | PDHFLTS |
| Renewable energy source selection/risk assessment | [37] | Interval-valued PLTS |
| | [47] | Multi-granular LD |
| Financial technologies selection | [54] | PLTS |
| Cloud-based ERP system selection | [7] | PLTS |
| Health-care waste disposal alternative selection | [33] | Multi-granular LD |
| Emergency decision-making | [22] | PLTS |
| | [42] | PLTS |
| | [44] | LD |
| Cloud vendor prioritization | [72] | PLTS |
| Quality function deployment | [78] | LD |

**(1) Applications in failure mode and effect analysis**

As a proactive risk management instrument, failure mode and effect analysis (FMEA) has been broadly utilized to recognize, evaluate, and eliminate failure modes of products, processes, systems and services [24]. Due to various subjective and objective conditions, it is often difficult for FMEA team members to provide precise values for the assessment of failure modes. Instead, they prefer to utilize linguistic labels to state their opinions regarding the risk of failure modes. Recently, the distributed linguistic representations have been adopted to model the uncertain opinions of FMEA team members. For example, Huang et al. [32] applied LDs to represent FMEA team members' risk evaluation information and employed an improved TODIM (an acronym in Portuguese of interactive and multicriteria decision making) method to determine the risk priority of failure modes. Ju et al. [34] proposed new approaches for heterogeneous linguistic FMEA with incomplete weight information as well as the idea to convert heterogeneous linguistic information to LDs.

Nie et al. [64] developed a hybrid risk evaluation model within the FMEA framework based on the use of multi-granular LDs and applied this model to supercritical water gasification system. Zhang et al. [110] proposed a consensus-based MAGDM approach for ordinal classification-based FMEA problem, in which FMEA participants provide their preferences in a linguistic way using LDs. Zhang et al. [113] integrated a consensus-reaching mechanism with bounded confidences into the FMEA framework and adopts ILDs to represent risk assessment information. Zhang et al. [111] presented the design of a PIS-based FMEA approach, in which members express their opinions over failure modes and risk factors using LDs.

**(2) Applications in hotel selection**

With the considerable development of tourism market, as well as the expansion of the e-commerce platform scale, increasing tourists often prefer to select tourism products such as services or hotels online. Thus, it is necessary to develop efficient decision support models for tourists to select tourism products. Liang et al. [43] developed decision support model for hotel selection based on sentiment analysis and LD-VIKOR method, where the text data are transformed into LDs via sentiment analysis. Nie et al. [62] proposed a scientific hotel selection model to assist tourists in choosing a satisfactory hotel and guiding hoteliers to gain competitive advantages in the e-tourism era, in which LDs are used to summarize and denote evaluation values under certain criteria associated with hotels. Peng et al. [67] developed an applicable hotel decision support model for tourists utilizing online reviews on TripAdvisor.com, in which PLTSs are introduced to summarize this information statistically by considering a great deal of review information associated with hotels posted by numerous tourists on TripAdvisor.com. Yu et al. [98] designed a mathematical model to select appropriate hotels on websites based on LDs. Zhang et al. [106] proposed a multi-stage MADM method based on online reviews for hotel selection considering the aspirations with different development speeds, in which LDs are used to deal with online reviews.

**(3) Applications in sustainability evaluation and other selection problems**

Water security sustainability plays an increasingly crucial role in maintaining the balance between industrialization, urbanization and sustainability. Nie et al. [63] established an observation data conversion standard and adopted LDs as information representation and developed a multistage decision support framework by combining several MADM techniques (such as, best-worst method, DEMATEL, and TOPSIS) for water security sustainability evaluation. With the accumulation of practical experience and the maturity of technology, constructed wetlands have gradually become multi-functional ecological systems. Luo et al. [52] proposed GDM approach for evaluating the sustainability of constructed wetlands with PLPRs.

The sustainable third-party reverse logistics provider selection, as the core of sustainable supply chain management, has become paramount in research nowadays. Luo and Li [53] utilized PDHFLTSs to deal with the situation in which decision makers may hesitate in a few linguistic terms and have different partiality towards each term in the actual evaluation process, and presented an MAGDM method by combining PDHFLTSs and MULTIMOORA for the sustainable third-party reverse logistics provider selection in the e-commerce express industry.

"No technology, no finance" has been the consensus in banking industry. Under the background of financial technology (Fintech), how to select an appropriate technology company to cooperate for the banks has become a key. Miao et al. [54] developed an MAGDM model with PLTSs for financial technologies selection. Krishankumar et al. [37] proposed a GDM framework for renewable energy source selection under interval-valued PLTSs.

Cloud-based enterprise resource planning (ERP) is a combination of standard ERP system and cloud flexibility. On the basis of extended probabilistic linguistic MULTIMOORA method and Choquet integral operator, Chen et al. [7] introduced an innovative two-step comparative method for the evaluation of cloud-based ERP systems. Ju et al. [33] presented a framework incorporating the evaluation based on distance from average solution method for selecting desirable health-care waste disposal alternative(s), in which multi-granular LDs are adopted by decision makers to assess the ratings of alternatives and subjective weights of criteria.

**(4) Applications in emergency decision making and prioritization events**

According to the characteristics of emergency management, Gao et al. [22] proposed an emergency decision support method by using the PLPR, and a case study about the emergency decision making in a petrochemical plant fire accident is conducted to illustrate the proposed method. Li and Wei [42] proposed an emergency decision-making method based on D-S evidence theory and PLTSs, and applied the proposed method to an actual mine accident.

With the tremendous growth of cloud vendors, cloud vendor prioritization is a complex decision-making problem. Sivagami et al. [72] proposed a scientific decision framework for Cloud Vendor prioritization under PLTSs context with unknown/partial weight information. Tian et al. [78] presented an improved quality function deployment for prioritizing service designs, in which multi-granular unbalanced linguistic term sets are used to capture evaluators' ratings to cope with vague information. Moreover, a unification method is proposed to convert multi-granular linguistic information into LDs.

## 4. Summary, critical discussion and challenges from the perspective of decision making and data science/XAI

In this section, we present the summary and critical discussion from a decision making perspective, and also present some ongoing challenges and future research directions from the perspective of data science and XAI.

## 4.1. Summary and critical discussion: A decision making perspective

The review about distributed linguistic representations is mainly summarized from a triple perspective:

(1) Origin of distributed linguistic representations. The distributed linguistic representations in the literature are mainly derived from: (i) extension of proportional linguistic 2-tuple representation, including LD, INLD, PLD, and ILD; and (ii) extending HFLTSs preferences, including PDHFLTS, PHFLTS, PLTS, HLD, and FLE. Particularly, these distributed linguistic representations can be formed in two ways: (i) distributed linguistic preference expressions in an individual context; and (ii) information fusion of linguistic terms or HFLTSs in a group context.

(2) Taxonomy of LD and its variants. We present the taxonomy of distributed linguistic representations. The LD is the first attempt to model the distributed linguistic representations among the decision-making community, and the follow-up notions and concepts are mathematically consistent with LD in some sense: FLE is the generalization of all the distributed linguistic representations reviewed; LD, PDHFLTS, PHFLTS, PLTS, PLD, and ILD are special HLDs; PLTS and ILD are mathematically consistent with PLD; PHFLTS is mathematically consistent with LD; PDHFLTS is the special case of LD; and INLD is a generalization of LD.

(3) Applications in decision making. We review different applications of distributed linguistic representations in decision making from four aspects. (i) Distributed linguistic distance measurements and aggregation methods. Most of them are based on classical distance measurements and aggregation operators. Moreover, optimization methods are also developed for dealing with accuracy problems in some decision contexts. (ii) Distributed linguistic pairwise comparison methods, including three types of distributed linguistic preference relations (i.e., DLPR, PLPR, and FLPR). (iii) Distributed linguistic MADM approaches. Several researchers revisited this issue based on classical MADM methods in distributed linguistic contexts. (iv) Real-life applications, in which participants involved prefer to adopt distributed linguistic representations to express their preferences to deal with practical decision problems.

Distributed linguistic representations have been widely studied to model the uncertainty and complexity of preference information in decision making, but there still exist some limitations to be further highlighted:

(1) Although distributed linguistic representations have been analyzed from multiple aspects, it still lacks a systematic research on the transformations among them. Relevant discussions on transformations from an axiom-based perspective are of significant necessity, and rational minimum information-loss based transformation models are needed to be designed as well. Notably, normalization methods have been proposed to transform PLTSs with incompleteness to LDs, but they are questionable. Thus, reasonable normalization methods from PLTSs to LDs are worth studying.

(2) In distributed linguistic representations many similar concepts have been put forward, and there is much parallel research in decision-making being undertaken, which lead to lots of repeated discussions and confusion of concepts. Therefore, it is necessary to focus on the original one and undertake valuable research and comparisons. Particularly, the FLE will be a potential tool in distributed linguistic representations to form a unified framework.

## 4.2. Challenges in data science and artificial intelligence: An explainable linguistic approach

The development of data science and artificial intelligence, particularly the efforts currently being made in the area of XAI [4], has provided enormous opportunities as well as arising challenges. How to fix the idea of XAI, and the opportunity to use linguistic information for representing the cognitive complex expressions from decision makers/experts or data driven approaches getting linguistic labels as knowledge, is needed to be addressed.

**(1) Natural language processing (NLP) method.** NLP is a significant artificial intelligence-based tool to process information linguistically. The potential in using NLP approaches in decision making has been partially shown in the literature [124], and the research on opinion mining from user data is being undertaken, such as accurate recognition of specific behaviors [5], credit risk assessments [61], etc. The advent of data science has brought the chance that NLP-based techniques can be used to handle a wider range of linguistic data [27], [36]. Particular attention should be paid to the data driven decision making using distributed linguistic representations as a representation approach for getting the information, experts' assessments from the data, and using NLP approaches for getting linguistic assessments from linguistic opinions. Essentially, challenges under this perspective rely on how to apply these NLP-based techniques (e.g., sentiment analysis) as well as data-driven technologies to analyze decision makers' distributed linguistic data, and emphasize on the necessity to figure out the characteristics of complexity of distributed linguistic information, and the difficulty to effectively handle these complexity with data-driven NLP tools.

**(2) Data fusion with data-driven and/or artificial intelligent approaches.** In the extent reviewed literature about the fusion methods of distributed linguistic information in decision making problems, the popular methods are based on the aggregation operators and mathematical optimization modelling. The development of mass media and internet technologies has witnessed more and more direct or indirect participation from different groups in the process of decision making. Complex distributed linguistic preferences are provided by decision makers with varying social backgrounds, self-confidence levels, knowledge structures, etc., and some key elements begin to appear in the decision making at a large scale [15], such as interactions with behaviors among decision makers/experts [6], [65], [93].

The challenges in this perspective centralize on coping with enormous amounts of distributed linguistic information from disparate data sources with data-driven and/or intelligent approaches. The analysis or processing of such information is no longer using traditional decision support tools or simple aggregation techniques. In contrast to the black models provided by the neural network based approaches, whose utility in data analysis have been limited because of the interpretation difficulty, linguistic based models with explainability are much more suitable. How to fuse these data and how to acquire significant insights from these data are still the future research needed to be investigated.

(3) **Data-driven preference learning methods.** Preference learning is a new research field of the intersection of machine learning and decision making. The preference learning mainly focuses on the analysis of individual and group characteristics, and modelling group/multi-attribute preference learning functions by learning historical data [2], [13]. It is worth pointing out that the advent of data science and artificial intelligence highlights the importance of preference learning methodology driven by large scale data sets [68]. The potential of applying data-driven preference learning methods in linguistic decision making is still under development. For example, how to learn the word encoding (semantic analysis), which is the core of CW, and model PIS-based semantic learning from distributed linguistic data at a large scale, is a problem worthy of further study.

Recent studies [1], [108] suggested that deep learning approaches are playing vital roles in parameter estimating and parameter settings for correlated features of decision models. Thereby, this learning method could be used to estimate optimal weights of multiple attributes in distributed linguistic MADM, and to determine suitable parameter settings for aggregation functions in the fusion process of distributed linguistic data. The potential of these machine learning methods can be detected in various practical applications, such as financial risk analysis [35], online social recommendation [122], etc.

(4) **Social network analysis (SNA) applications.** With the development of social media and communication technologies, it is easier to obtain the access to information about social networks, which makes interactions among decision makers become increasingly common. As a powerful tool, SNA has been investigated to support the decision process from many aspects such as opinion dynamics [20], trust/distrust relationships [79], [117], GDM [19], and LSDM [15], in which relationships among decision makers are modelled in a social network. Combining the distributed trust/distrust relationships among decision makers analyzed by SNA, the preference characteristics of decision makers are identified, which can further support the management of decision makers' behaviors and the approaches for generating recommendations to improve the efficiency of decision process. It would also be interesting to apply these SNA-based decision models into real-life decision problems. For example, in the hotel selection driven by online textual reviews [62], customers' sentiments can be analyzed to extract both preferences and distributed trust relationship information as an input for decision models.

(5) **Online customer reviews application.** Mass media and platforms have been vested with significant power in society. For example, it is noted that the spread of e-commerce platforms and mobile apps have improved customers' ability to assess products online. Online reviews on various e-commerce platforms have become an important part of the electronic word of mouth and the important references for potential customers to make decisions [48]. These online reviews/data have significant distribution characteristics and effective analysis or processing of these data can help enterprises/government understand the consumer preference much better, and make rational decision accordingly.

In future research, a comprehensive analysis of the complexity of consumption data sources is needed. In particular, NLP-based, emotion recognition-based [121], SNA-based, and PIS-based approaches [45] are powerful tools to better and accurately understand the sociality and individuality of customers through these distributed linguistic data.

## 5. Conclusion

We review the distributed linguistic representations in decision making from a triple perspective of taxonomy, key elements and applications, and ongoing challenges. Specifically, we analyze the origin of existing distributed linguistic representations in the literature, which are classified into two types: extension of proportional linguistic 2-tuple representation, and HFLTS based extended representation. Particularly, these distributed linguistic representations can be formed in two ways: linguistic preference expressions in an individual context, and information fusion of linguistic terms or HFLTSs in a group context. Furthermore, we summarize the key elements and applications of distributed linguistic representations in various decision problems, including distributed linguistic distance measurements and aggregation methods; distributed linguistic preference relations; distributed linguistic MADM; and some real-life applications. Finally, we critically discuss the concept confusion issues of distributed linguistic representations, and propose the challenges and future directions from the perspective of data science and XAI.

**Acknowledgments**

We thank Dr. Cuiping Wei (Yangzhou University) and Dr. Zhen-Song Chen (Wuhan University) for their valuable suggestions to improve this review. This work was supported by the grants (Nos. 71971039, 71421001, 71910107002, 71771037, 71874023, and 71871149) from NSF of China, and the grants (Nos. sksyl201705 and 2018hhs-58) from Sichuan University.